\documentclass[10pt,twocolumn,letterpaper]{article}

\usepackage{cvpr}
\usepackage{times}
\usepackage{epsfig}
\usepackage{graphicx}
\usepackage{amsmath}
\usepackage[normalem]{ulem}
\usepackage{amssymb}
\usepackage{enumitem}
\usepackage{xcolor}
\usepackage{authblk}
\newenvironment{packed_itemize}{
\begin{enumerate}[leftmargin=*]
  \setlength{\itemsep}{0.6pt}
  \setlength{\parskip}{0pt}
  \setlength{\parsep}{0pt}
}{\end{enumerate}}

\newenvironment{packed_enum}{
\begin{itemize}[leftmargin=*]
  \setlength{\itemsep}{0.6pt}
  \setlength{\parskip}{0pt}
  \setlength{\parsep}{0pt}
}{\end{itemize}}
\DeclareMathOperator*{\argmax}{arg \, max}
\usepackage[pagebackref=true,breaklinks=true,letterpaper=true,colorlinks,bookmarks=false]{hyperref}

 \cvprfinalcopy 

\def\cvprPaperID{6220} 

\ifcvprfinal\fi
\begin{document}
\title{Don't Judge an Object by Its Context: Learning to Overcome Contextual Bias}


\makeatletter
\renewcommand\AB@affilsepx{, \protect\Affilfont}
\makeatother
\renewcommand{\Authands}{ , }

\author[1]{Krishna Kumar Singh}
\author[2]{ Dhruv Mahajan}
\author[2,3]{Kristen Grauman}
\author[1]{Yong Jae Lee}
\author[2]{Matt Feiszli}
\author[2]{ \\Deepti Ghadiyaram}
\affil[1]{University of California, Davis}
\affil[2]{Facebook AI} 
\affil[3]{University of Texas at Austin}

\newcommand{\biasedClassNoMath}{\mathrm{b}}
\newcommand{\biasedClass}{$\biasedClassNoMath$\xspace}
\newcommand{\cooccurClassNoMath}{\mathrm{c}}
\newcommand{\cooccurClass}{$\cooccurClassNoMath$\xspace}
\newcommand{\biasedClassWithINoMath}{\mathrm{{b}_{j}}}
\newcommand{\biasedClassWithI}{$\biasedClassWithINoMath$\xspace}
\newcommand{\cooccurClassWithINoMath}{\mathrm{{c}_{j}}}
\newcommand{\cooccurClassWithI}{$\cooccurClassWithINoMath$\xspace}
\newcommand{\biasedClassWithJNoMath}{\mathrm{{b}_{j}}}
\newcommand{\biasedClassWithJ}{$\biasedClassWithJNoMath$\xspace}
\newcommand{\cooccurClassWithJNoMath}{\mathrm{{c}_{j}}}
\newcommand{\cooccurClassWithJ}{$\cooccurClassWithJNoMath$\xspace}
\newcommand{\biasedSetPairNoMath}{\mathrm{\mathbb{S}}}
\newcommand{\biasedSetPair}{$\biasedSetPairNoMath$\xspace}
\newcommand{\meanFeatureNoMath}{\mathrm{\bar{{x}_{s}}
}}
\newcommand{\alphaMinNoMath}{\mathrm{\alpha_{min}}}
\newcommand{\alphaMin}{$\alphaMinNoMath$\xspace}

\newcommand{\meanFeature}{$\meanFeatureNoMath$\xspace}
\newcommand{\objectWeightNoMath}{\mathrm{W_o}}
\newcommand{\objectWeight}{$\objectWeightNoMath$\xspace}
\newcommand{\contextWeightNoMath}{\mathrm{W_s}}
\newcommand{\contextWeight}{$\contextWeightNoMath$\xspace}
\newcommand{\fullFeatureNoMath}{\mathrm{x}}
\newcommand{\fullFeature}{$\fullFeatureNoMath$\xspace}
\newcommand{\biasedFeatNoMath}{\mathrm{x_{o}}}
\newcommand{\biasedFeat}{$\biasedFeatNoMath$\xspace}
\newcommand{\cooccurFeatNoMath}{\mathrm{x_{s}}}
\newcommand{\cooccurFeat}{$\cooccurFeatNoMath$\xspace}
\newcommand{\cocoStuff}{COCO-Stuff\xspace}
\newcommand{\deepFashion}{DeepFashion\xspace}
\newcommand{\awa}{Animals with Attributes\xspace}
\newcommand{\unrel}{UnRel\xspace}
\newcommand{\standard}{\textit{standard}\xspace}
\newcommand{\ourCam}{\textit{ours-CAM}\xspace}
\newcommand{\ourFeat}{\textit{ours-feature-split}\xspace}
\newcommand{\dineshDecorrelate}{\textit{attribute decorrelation}\xspace}
\newcommand{\classBalance}{\textit{class balancing loss}\xspace}
\newcommand{\splitBias}{\textit{split biased}\xspace}
\newcommand{\rmCoLabels}{\textit{remove co-occur labels}\xspace}
\newcommand{\rmCoImages}{\textit{remove co-occur images}\xspace}
\newcommand{\wtLoss}{\textit{weighted loss}\xspace}
\newcommand{\negPenalty}{\textit{negative penalty}\xspace}
\maketitle

\begin{abstract}
Existing models often leverage co-occurrences between objects and their context to improve recognition accuracy. However, strongly relying on context risks a model's generalizability, especially when typical co-occurrence patterns are absent. This work focuses on addressing such contextual biases to improve the robustness of the learnt feature representations. Our goal is to accurately recognize a category in the absence of its context, without compromising on performance when it co-occurs with context.
Our key idea is to decorrelate feature representations of a category from its co-occurring context. We achieve this by learning a feature subspace that explicitly represents categories occurring in the absence of context along side a joint feature subspace that represents both categories and context. Our very simple yet effective method is extensible to two multi-label tasks -- object and attribute classification. On $4$ challenging datasets, we demonstrate the effectiveness of our method in reducing contextual bias.

\end{abstract}

\section{Introduction}
Visual context serves as a valuable auxiliary cue for the human visual system for scene interpretation and object recognition~\cite{biederman1982scene}. Context can either be a co-occurrence of objects and scenes (e.g., ``boat'' is often present in ``outdoor waters'') or of two or more objects in a given scene (e.g., ``skis'' often co-occur with a ``skier''). Context becomes especially crucial for our visual system when the visual signal is ambiguous or incomplete (e.g., due to occlusion, viewpoint of the scene capture, etc.). Past research explicitly models context and shows benefits on standard visual tasks such as classification~\cite{tang-cvpr15} and detection~\cite{divvala-cvpr09,barnea-cvpr19}. Meanwhile, convolution networks by design implicitly capture context.
\begin{figure}[t!]
 \centering
 \includegraphics[width=0.5\textwidth]{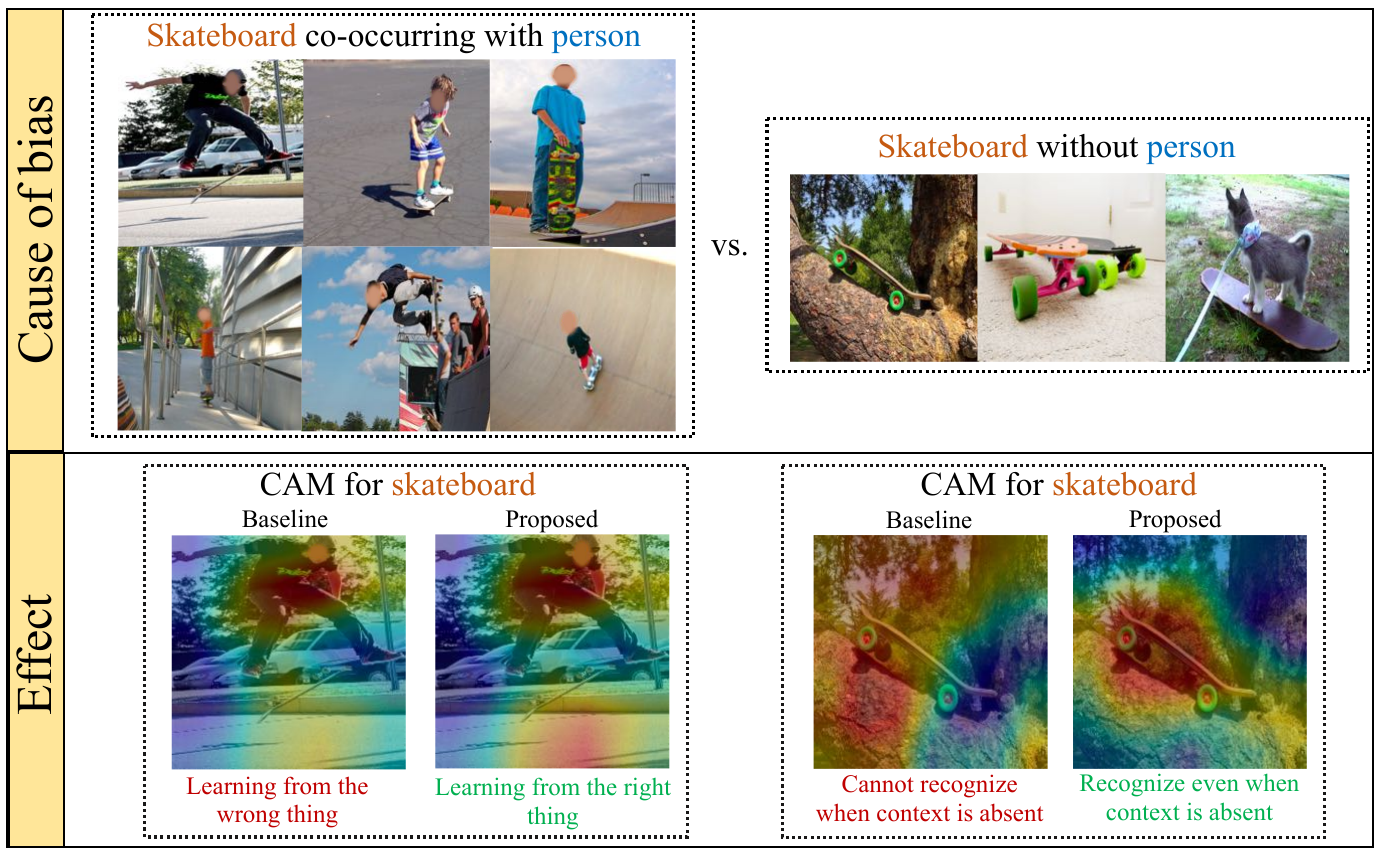}
 \vspace{-0.15in}
 \caption{\footnotesize{\textbf{Top (cause of contextual bias)}: Sample training images of the category ``skateboard''. Notice how it very often co-occurs with ``person'' and how all images are captured from similar viewpoints. In the rare cases where skateboard occurs exclusively, there is higher viewpoint variance. \textbf{Bottom (effect of such bias)}: Such data skew causes a typical classifier to rely on ``person'' to classify ``skateboard'' and worse, unable to recognize skateboard when person is absent. Our proposed approach overcomes such contextual bias by learning feature representations that decorrelate the category from its context.}}
 \vspace{-0.25in}
 \label{fig:teaser}
\end{figure}

Deep networks rely on the availability of large-scale annotated datasets~\cite{lin-eccv14, deng_cvpr09} for training. As highlighted in~\cite{torralba-cvpr11, tommasi-2017}, despite the best efforts of its creators, most prominent vision datasets are afflicted with several forms of \textit{biases}. Let us consider an object category ``microwave.'' A significant portion of images belonging to this category are likely to be captured in kitchen environments, where other objects such as ``refrigerator,'' ``kitchen sink,'' and ``oven'' frequently co-occur. This may inadvertently induce \textit{contextual bias} in these datasets, which would consequently seep into models trained on them. Specifically, in the process of learning features that separate positive and negative instances in such a (biased) training dataset, a deep discriminative model can very often also strongly capture the context co-occurring with the category of interest. This issue is exacerbated in a setting where we do not have explicit location annotations (e.g., bounding boxes and segmentation masks) of such biased categories, and a model being trained has to rely solely on image-level annotations to perform multi-label classification. Having a model \textit{implicitly} learn to localize such context-biased categories in the absence of location annotations is challenging.

Does it even matter if a model inadvertently learns such correlations? We believe this can cause problems on two fronts: (1) failing to identify ``microwave'' in a different context such as an ``outdoor'' scene or in the \textit{absence} of ``refrigerator'' and (2) hallucinating ``refrigerator'' even in an indoor kitchen scene containing only ``microwave.'' The issue of co-occurring bias is also prevalent in visual attributes~\cite{liu-cvpr16, xian-tpami18}. For example, in the Deep Fashion dataset~\cite{liu-cvpr16}, the attribute ``trapeze'' strongly co-occurs with ``striped.'' This results in a less credible classifier that has a hard time recognizing ``trapeze'' in clothes with ``floral.'' Recent research has identified far more serious mistakes made by trained models due to inherent biases in both language and vision datasets -- learning correlations between ethnicity and certain sport activities~\cite{stock2018convnets}, gender and profession~\cite{bolukbasi2016man, hendricks2018women, zhao2017men}, and age and gender of celebrities~\cite{alvi-eccv18}. Such grave confusion caused due to biases in the data impedes the deployment of these models in real-world applications. 

Given these issues, our goal is to train an unbiased visual classifier that can accurately recognize a category both in the presence and absence of its context. Specifically, given two categories with a strong co-occurring bias, our aim is to accurately recognize them when either one occurs \textit{exclusively}, and  at the same time not hurt the performance when they \textit{co-occur}. To this end, we propose two key ideas. First, we hypothesize that a network should learn about a category by relying more on its corresponding pixel regions than those of its context. Since we only have class labels, we use class activation maps (CAM)~\cite{zhou-cvpr16}  as ``weak'' location annotations and minimize their mutual spatial overlap. 

Building on this, we devise a second method that learns feature representations to decorrelate a category from its context. While the entire feature space learned by the network jointly represents category and context, we explicitly carve out a subspace to represent categories that occur away from typical context. We learn this feature subspace only from training instances where a biased category occurs in the absence of its context. In all other cases, the model \textit{should} also leverage context and thus the entire feature space. At test time, we make no such distinction and the entire feature space is equally leveraged. Therefore, in the example from Fig.~\ref{fig:teaser}, our goal is to learn a feature subspace to represent ``skateboard'' while the entire feature space jointly represents ``skateboard'' and ``person.''

Through extensive evaluation, we demonstrate significant performance gains for the hard cases where a category occurs away from its typical context.  Crucially, we show that our framework does not adversely effect recognition performance when categories and context co-occur.
To summarize, we make the following contributions:
\vspace{-0.07in}
\begin{packed_enum}
 \item With an aim to teach the network to ``learn from the right thing,'' we propose a method that minimizes the overlap between the class activation maps (CAM) of the co-occurring categories (Sec.~\ref{sec:cam}).
 \item Building on the insights from the CAM-based method, we propose a second method that learns feature representations that decorrelate context from category (Sec.~\ref{sec:framework}).
 \item We apply both methods on two tasks: object and attribute classification, and $4$ datasets, and achieve significant boosts over strong baselines for the hard cases where a category occurs away from its typical context (Sec.~\ref{sec:experiments}). 
\end{packed_enum}

\section{Related work}
\noindent \textbf{Addressing biases:}
Prior work~\cite{torralba-cvpr11, khosla-eccv12, van-arxiv16, tommasi-2017} has shown that existing datasets suffer from bias and are not perfectly representative of the real world. Hence, a model trained on such data will have difficulty generalizing to non-biased cases. Attempts to reduce dataset bias include domain adaptation techniques~\cite{csurka-arxiv17} and data re-sampling~\cite{chawla2002smote,li-cvpr19}, e.g., so that minority class instances are better represented.  One limitation of data re-sampling is that it can involve reducing the dataset, leading to sub-optimal models.  Recent adversarial learning approaches~\cite{alvi-eccv18, kim-cvpr19} try to mitigate bias from the learned feature representations while optimizing performance for the task at hand (e.g., removing gender bias while classifying age). However, these methods would not be directly applicable for mitigating \emph{contextual} bias, as context (the bias factor) can still be useful for recognition---so it cannot be simply removed. Others study various forms of bias in the context of image captioning (e.g., gender bias)~\cite{hendricks2018women}, image classification (e.g., ethnicity bias)~\cite{stock2018convnets}, and object recognition (e.g., socio-economic bias)~\cite{vries-cvprw19}.  Overall, \emph{contextual} bias in visual recognition remains relatively under explored.

\noindent \textbf{Co-occurring-bias:}
Contextual bias is a well-studied problem in the field of natural language processing ~\cite{recasens-acl13,sun-arxiv19}, however, it is much less studied in the computer vision community.  In vision, most efforts consider context as a useful cue~\cite{divvala-cvpr09,barnea-cvpr19}. A few efforts have shown that a recognition model will fail to recognize an object without its co-occurring context, but do not propose a solution~\cite{choi-prl12,rosenfeld-arxiv18}.

A recent method reduces contextual bias in video action recognition~\cite{wang-cvpr18}, but it relies on temporal information and thus cannot be applied to the image recognition problems we tackle in this work.  A pre-deep learning approach~\cite{jayaraman-cvpr14} reduces the correlation (bias) between visual attributes by leveraging additional knowledge in the form of semantic groupings of attributes. Recently~\cite{zhu-cvpr19} tried to reduce contextual bias for object detection by learning focused foreground features, but they require expensive bounding-box annotations. In contrast, our deep learning approach does not require any additional supervision apart from the object/attribute class labels.  Most importantly, to our knowledge, there is no prior work focusing on mitigating contextual bias for object classification as we do in this paper.
\\ \noindent \textbf{Relation to few-shot learning:}
Lastly, contextual bias could also be formulated as a few-shot~\cite{snell-nips2017,karlinsky-cvpr19,alfassy-cvpr19} or class imbalance~\cite{elkanfoundations,cui-cvpr19} problem, since images in which objects appear without their usual co-occurring context (e.g., keyboard without a mouse next to it) are relatively rare. However, treating such rare (exclusive) images as a separate class or simply assigning them higher weight can be sub-optimal, as we show in our experiments.

\section{Problem setup} \label{sec:formulation}
Our method operates on the premise that the training data distribution corresponding to a few categories suffers from co-occurring bias. We henceforth refer to them as \textit{biased categories}. We make no such assumptions about the test data distribution. For example, \cocoStuff~\cite{caesar-cvpr18} has $2209$ images where ``ski'' co-occurs with ``person,'' but only has $29$ images where ``ski'' occurs without ``person.'' A model trained on such skewed data may fail to recognize when ``ski'' occurs in isolation. Our goal is to learn a feature space that is robust to such training data biases. In particular, given a (presumably) unbiased test dataset, our goal is to (1) correctly identify ``ski'' when it occurs in isolation and (2) not lose performance when ``ski'' co-occurs with ``person.'' A key aspect of our approach is to identify most biased categories for a given dataset, which we describe next.
\subsection{Identifying biased categories} \label{sec:bias_formulation}
\begin{figure}[t!]
 \centering
 \includegraphics[width=0.2\textwidth]{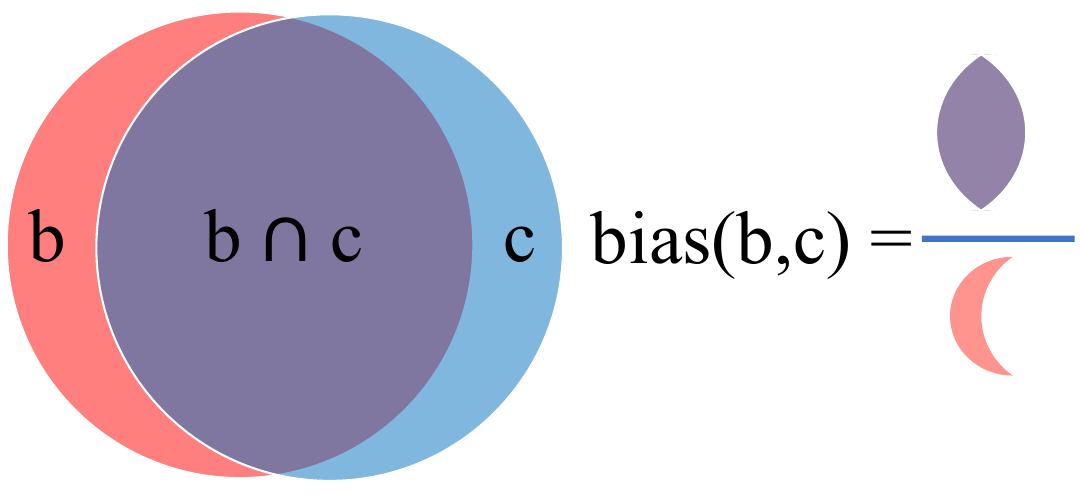}
 \caption{\footnotesize{\textbf{Quantifying bias} in \biasedClass due to its high co-occurrence with \cooccurClass. 
}}
 \label{fig:bias}
 \vspace{-0.15in}
\end{figure}
Suppose we are learning a classifier on a multi-label training dataset with a vocabulary of $\mathrm{M}$ categories. Only a few of these categories suffer from context\footnote{Throughout, we use context and co-occurring interchangeably.} bias; thus, a key aspect of our approach is to find this set of $\mathrm{K}$ category pairs $\biasedSetPairNoMath =\{($\biasedClassWithJ $,$ \cooccurClassWithJ $)\}$, where $\mathrm{0 \le j < K}$, which suffer the most from co-occurring bias\footnote{Although we consider pairs of co-occurring categories throughout, the proposed method is extensible for any number of co-occurring categories.}. Henceforth, \biasedClassWithJ (e.g. ``ski'') denotes a class which is most biased with \cooccurClassWithJ (e.g. ``person'') due to its high co-occurrence. \\
\textbf{Intuition:} While there are several ways to construct \biasedSetPair, our method is built on the following intuition: a given category \biasedClass is most biased by \cooccurClass if (1) the prediction probability of \biasedClass drops significantly in the \textit{absence} of \cooccurClass and (2) \biasedClass co-occurs frequently with \cooccurClass. 

We now define our method to identify \cooccurClass for a given \biasedClass. For a given category $\mathrm{z}$, let  $\mathrm{\mathbb{I}_{\biasedClassNoMath} \cap \mathbb{I}_{z}}$ and $\mathrm{\mathbb{I}_{\biasedClassNoMath} \setminus \mathbb{I}_{z}}$ denote sets of images where \biasedClass occurs with and without $\mathrm{z}$ respectively. Let $\mathrm{\hat{p}(i,\biasedClassNoMath)}$ denote the prediction probability of an image $\mathrm{i}$ for a category \biasedClass obtained from training a standard multi-label classifier. We quantify the extent of $bias$ between \biasedClass and $\mathrm{z}$ as follows: 
\begin{equation} \label{eqn:bias}
\mathrm{
    bias(\biasedClassNoMath,z) =  \frac{\frac{1}{|\mathbb{I}_{\biasedClassNoMath} \cap \mathbb{I}_{z}|}\sum\limits_{I \in \mathbb{I}_{\biasedClassNoMath} \cap \mathbb{I}_{z}} \hat{p}(i,\biasedClassNoMath)}{\frac{1}{|\mathbb{I}_{\biasedClassNoMath} \setminus \mathbb{I}_{z}|}\sum\limits_{I \in \mathbb{I}_{\biasedClassNoMath} \setminus \mathbb{I}_{z}} \hat{p}(i,\biasedClassNoMath)},
    }
\end{equation}
where $|.|$ denotes cardinality of a set. Eq (\ref{eqn:bias}) measures the ratio of average prediction probabilities of the category \biasedClass when it occurs with and without $\mathrm{z}$ (see Fig.~\ref{fig:bias}). A higher value indicates a higher dependency of \biasedClass on $\mathrm{z}$. We determine \cooccurClass as follows:
 \begin{equation}
 \mathrm{
     \cooccurClassNoMath = \argmax_z bias(\biasedClassNoMath,z)
     }
 \end{equation}
i.e., for each \biasedClass, we identify a category \cooccurClass that (i) yields the highest value of $\mathrm{bias}$ and (ii) co-occurs at least $10 - 20\%$ times (see Sec.~\ref{sec:train_setup}) with \biasedClass. We then construct \biasedSetPair with $\mathrm{K}$ most biased category pairs. We note that the above formulation is directional, i.e., it only captures the biases in \biasedClass caused due to \cooccurClass. For instance, $\mathrm{bias(ski, person)}$ only captures bias in ``ski'' due to ``person'' but not vice-versa.

We next propose two methods to combat co-occurring bias in the training data. The input to both methods is (1) training images and their associated weak (multiple) category labels and (2) the set \biasedSetPair composed of the $\mathrm{K}$ most biased category pairs (identified from Eq. (\ref{eqn:bias})).
We stress that training images have only weak labels stating which categories are present; they have no spatial annotations to say \textit{where} in the image each category is.

\section{Approach}
Our first method relies on class activation maps (CAM) as ``weak'' automatically inferred location annotations and minimizes their spatial overlap between biased categories (Sec.~\ref{sec:cam}). Building on the observations from this CAM-based approach, we propose a second method which learns a feature space by encouraging context sharing when a biased category co-occurs with context while suppressing context when it occurs in isolation (Sec.~\ref{sec:framework}).
\subsection{CAM as ``weak'' location annotation}\label{sec:cam}
\begin{figure}[t!]
 \centering
 \includegraphics[width=0.4\textwidth]{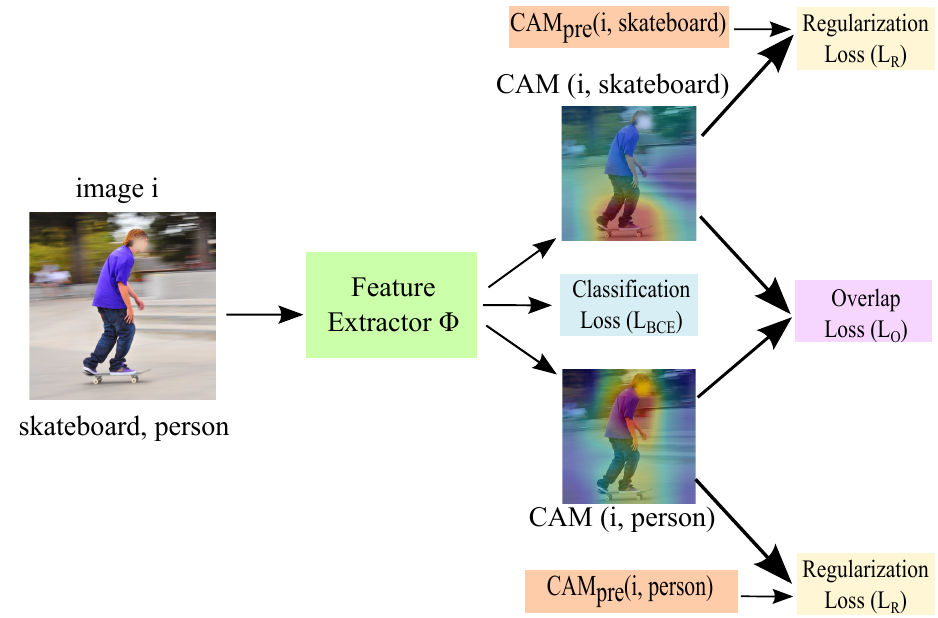}
 \caption{\footnotesize{\textbf{Our CAM-based approach} operates on category labels and requires no ground-truth location annotations. Instead, we leverage CAMs as weak location annotations and propose to minimize the mutual overlap between a biased category and its co-occurring context.}}
 \label{fig:approach_cam}
\vspace{-0.18in}
\end{figure}

Our method operates on the following premise: as \biasedClass almost always co-occurs with \cooccurClass, the network may learn to inadvertently rely on pixels corresponding to \cooccurClass to predict \biasedClass. This is particularly problematic when the network is tested on images where \biasedClass occurs in the absence of \cooccurClass. We hypothesize that one way to overcome this issue is to \textit{explicitly} force the network to rely less on \cooccurClass's pixel regions, \textit{without} using location annotations. While this may not succeed for occluding pairs like ``person'' and ``shirt,'' it seems like a natural constraint for spatially-distinct categories like ``person'' and ``skateboard.''

\noindent\textbf{Class Activation Maps:} To this end, we propose to use class activation maps (CAM)~\cite{zhou-cvpr16} as a proxy for object localization information. For a given image $\mathrm{i}$ and class $\mathrm{r}$, $\mathrm{CAM(i,r)}$ indicates the discriminative image regions used by a deep network to identify $\mathrm{r}$. Specifically, the final convolutional layer ($\mathrm{conv_{f}}$) of any typical network is followed by a global pooling and a fully connected ($\mathrm{fc}$) layer which predicts a score for class $\mathrm{r}$ in image $i$. $\mathrm{CAM(i,r)}$ is generated by \textit{projecting back} the 
weights of the $\mathrm{fc}$ layer for $\mathrm{r}$ on $\mathrm{conv_{f}}$ and computing a weighted average of the feature maps. Though CAMs are typically used as a visualization technique, in this work, we also use them to reduce contextual bias as we describe next.\\
\noindent\textbf{Formulation:}
In our setup, for each biased category pair $(\biasedClassNoMath, \cooccurClassNoMath)$ in \biasedSetPair (defined in Sec.~\ref{sec:bias_formulation}), we enforce minimal overlap of their CAMs via the loss function:
\begin{equation} \label{eqn:cam_loss}
\mathrm{L_{O}=\sum\limits_{i \in \mathbb{I}_{\biasedClassNoMath} \cap \mathbb{I}_{\cooccurClassNoMath}} CAM(i,\biasedClassNoMath) \odot CAM(i,\cooccurClassNoMath)}
\end{equation}

CAM offers two nice properties: (1) it is learned only through class labels without requiring any annotation effort and (2) it is fully differentiable, and thus can be integrated in an end-to-end network during training. 

Ideally, Eq (\ref{eqn:cam_loss}) should learn to reduce the spatial overlap between co-occurring categories, without hurting the classification performance. However, while attempting to minimize overlap, Eq (\ref{eqn:cam_loss}) could also lead to a trivial solution where the CAMs of \biasedClass and \cooccurClass drift apart from their actual pixel regions. To prevent this without strongly-supervised spatial annotations, we introduce a regularization term $\mathrm{L_{R}}$. Specifically, we pre-train a separate network (offline) for the standard classification task and generate $\mathrm{CAM_{pre}}$ from it for \biasedClass and \cooccurClass. We then \textit{ground} the CAMs of each category to be closer to its pixel regions predicted from $\mathrm{CAM_{pre}}$. $\mathrm{L_{R}}$ is thus defined as follows:
\begin{small}
\begin{equation}
\begin{split}
\mathrm{
L_{R} = \sum\limits_{i \in \mathbb{I}_{\biasedClassNoMath} \cap \mathbb{I}_{\cooccurClassNoMath}} |CAM_{pre}(i,\biasedClassNoMath)-CAM(i,\biasedClassNoMath)| + } \\ 
\mathrm{|CAM_{pre}(i,\cooccurClassNoMath)-CAM(i,\cooccurClassNoMath)|}
\end{split}
\end{equation}
\end{small}
We use a standard binary cross-entropy loss ($\mathrm{L_{BCE}}$) for the task of multi-label classification. Thus, our final loss becomes:
\begin{equation}
\mathrm{L_{CAM}= \lambda_{1} L_{O} + \lambda_{2} L_{R} + L_{BCE},}
\end{equation}
Fig.~\ref{fig:approach_cam} for the entire approach. As we show in results (Sec.~\ref{sec:experiments}), our CAM-based method successfully learns to rely more on the biased category's pixel regions thereby improving recognition performance. Our method yields large gains when a biased category occurs in the absence of its typical context. However, it sometimes hurts performance when biased category co-occurs with context (discussed later in Fig.~\ref{fig:cam_feat}). One reason could be that the pixel regions surrounding the co-occurring category also offer useful complementary information for recognizing the biased category. By discouraging mutual spatial overlap, CAM-based approach may not be able to leverage this information. This key insight led to the formulation of our next approach, which splits the feature space into two and separately represents context and category, while posing no constraints on their spatial extents.

\subsection{Feature splitting and selective context suppression}
\label{sec:framework}
\begin{figure*}[t!]
 \centering
 \includegraphics[width=0.8\textwidth]{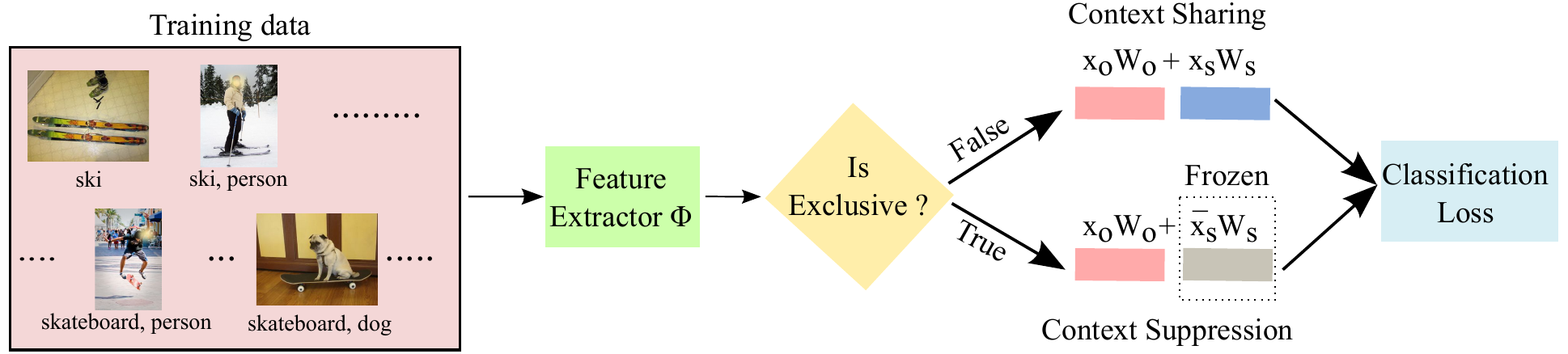}
 \caption{\footnotesize{\textbf{Our feature splitting approach} where images and their associated category labels are provided as input. During training, we split the feature space into two equal sub spaces: \biasedFeat and \cooccurFeat. If a training instance has a biased category occurring in the absence of context, we suppress \cooccurFeat (no back-prop), forcing the model to leverage \biasedFeat. In all other scenarios, \biasedFeat and \cooccurFeat are treated equally. At inference, the entire feature space is equally leveraged.}}
 \label{fig:approach}
 \vspace{-0.2in}
\end{figure*}

Rather than optimizing CAMs, we propose to learn a feature space that is robust to the inherent co-occurring biases in the training data. We observe that cases when a biased category co-occurs with context are often visually distinct from those where it occurs exclusively (see Fig.~\ref{fig:teaser}). This motivates us to learn a dedicated feature (sub) space to represent biased categories occurring away from their typical context. While the entire feature space learned by the model jointly represents context and category, this dedicated subspace should decouple the representations of a category from its context. We learn this feature subspace only from training instances where biased categories occur in the absence of their typical context. These modifications only affect training; at inference time the architecture is identical to the standard model.

\noindent\textbf{Formulation:} 
Given a deep neural network $\mathrm{\phi}$, let \fullFeature denote the $\mathrm{D}$-dimensional output of the final pooling layer just before the fully-connected layer ($\mathrm{fc}$). Let the weight matrix associated with $\mathrm{fc}$ layer be $\mathrm{W \in R^{D \times M}}$, where $\mathrm{M}$ denotes the number of categories in a given multi-label dataset. The predicted scores inferred by a classifier (ignoring the bias term) are
\begin{equation}
\mathrm{
\hat{y} = \mathrm{W^{T}x}.
}
\end{equation}
Because we wish to separate the feature representations of a category from its context, we (row-wise) split $\mathrm{W}$ randomly into two disjoint subsets: \objectWeight and \contextWeight, each of dimension $\frac{D}{2} \times M$. Consequently, \fullFeature is split into \biasedFeat and \cooccurFeat and the above equation can be rewritten as:
\begin{equation}
\mathrm{
\hat{y} = \mathrm{W^{T}_o x_o + W^{T}_s x_s}.
}
\end{equation}

In scenarios where a biased category occurs in the absence of its context, we want to \textit{enforce} the network to only rely on \objectWeight by suppressing \contextWeight. This step allows the network to explicitly capture the biased category-specific information when it occurs away from its context in \objectWeight. On the other hand, when a biased category co-occurs with its context, we want to \textit{encourage} the network to leverage both \objectWeight and \contextWeight. This would allow the network to jointly encode category and context in the full feature space. 

To achieve this, we make two minor modifications to a standard classifier when a biased category occurs away from its typical context. First, we 
disable back propagation through \contextWeight thereby forcing the network to learn only through \objectWeight. Second, we set \cooccurFeat to a constant value. We believe these two simple modifications allow us to suppress context in selective cases, i.e., when a biased category occurs away from its context. For instance, when \textit{ski} occurs in the absence of its typical context \textit{person}, our method suppresses \contextWeight thereby encouraging \objectWeight to encode its appearance; when \textit{ski} co-occurs with \textit{person}, both \objectWeight and \contextWeight are leveraged. 

In practice, we set \cooccurFeat $=$ \meanFeature, where \meanFeature is the average of \cooccurFeat over the last $10$ mini-batches, and allowed stabler training. Also, \meanFeature is a closer approximation to the range of values \cooccurFeat witnesses at test time. 

\noindent \textbf{Intuition behind weighted loss:} An underlying aspect of our method is that the biased categories occur very rarely in the absence of their context, making the training data distribution skewed (see Sec.~\ref{sec:formulation}). This is a problem since \objectWeight is learned solely from the (very few) samples with biased categories occurring in the absence of their typical context. We address this issue by associating a higher weight to such training samples. All other samples are weighed equally. Specifically, we define a weight $\alpha$ such that

\begin{equation} \small
\alpha = \begin{cases} \sqrt{\frac{|\mathbb{I}_{\biasedClassNoMath} \cap \mathbb{I}_{\cooccurClassNoMath}|}{|\mathbb{I}_{\biasedClassNoMath} \setminus \mathbb{I}_{\cooccurClassNoMath}|}}, &  \text{when \biasedClass occurs exclusively}  \\ \\ \hspace{0.2in} 1 & \text{otherwise} \end{cases}
\end{equation}

Thus, $\alpha$ is the ratio of the number of training instances where category occurs in the presence vs. absence of context. A higher value of $\alpha$ for a given biased category indicates more data skewness. \footnote{In practice, we ensure $\alpha$ is at least \alphaMin (a constant value $ > 1$) when \biasedClass occurs exclusively.}. 

Given ground-truth label $\mathrm{t}$ and sigmoid function $\sigma$, our weighted binary cross-entropy loss is defined as follows:
\begin{equation} \label{eqn:bce}
\mathrm{
L_{BCE}= -\alpha \left(t log(\sigma({\hat{y}})) + (1-t) log(1- \sigma({\hat{y}}))\right)
},
\end{equation}

Figure~\ref{fig:approach} illustrates the proposed method. While a standard classifier jointly encodes category and context, it fails to recognize biased categories occurring without context. By contrast, our approach splits the feature space and represents biased categories occurring without context in a dedicated subspace. As we will show in results, due to selective context suppression, this feature subspace successfully captures category-specific information. Furthermore, in the second subspace, our method effectively leverages context when available and jointly encodes it with category. 

As we show in results, leveraging context when available, distinguishes this method with the CAM-based method described in Sec.~\ref{sec:cam} and plays a key role in recognition performance. Further, while we selectively suppress context when a biased category occurs away from its context, the CAM-based method optimizes the mutual spatial overlap when a biased category co-occurs with context. 
We stress that both methods are applied only for the $\mathrm{K}$ biased category pairs; thus, misclassification loss for the other (non-biased) categories also plays an important role in learning. Finally, our method poses no constraints on the spatial extents of categories; thus, unlike our CAM-based approach, is extensible to attributes.
\subsection{Training setup}\label{sec:train_setup}

\noindent\textbf{Determining biased categories:}
For each category, we first identify other categories that occur frequently (at least $\mathrm{10\% - 20\%}$ times, based on the dataset). Next, we partition the training data into non-overlapping $80-20$ split. We train a standard multi-class classifier with BCE loss on the $\mathrm{80\%}$ split and compute $\mathrm{bias}$ (Eq. \ref{eqn:bias}) on the $\mathrm{20\%}$ split. While both methods proposed in this work can be applied to any number of biased category pairs, we found that setting $\mathrm{K = 20}$ (Sec.~\ref{sec:bias_formulation}) sufficiently captures biased categories in all the datasets we study here.

\noindent\textbf{Optimization:}
We follow a two-stage training procedure: in the first stage, we start with a pre-trained network as a backbone and fine-tune it on all categories of a given dataset. This step ensures that the network learns useful context cues for the target task. 
In the second stage, we fine-tune our network and separately apply the modified loss defined in each proposed method. In the CAM-based approach, we reduce spatial overlap between the $\mathrm{|K|}$  category pairs; in the feature splitting method, we  selectively suppresses context when the $\mathrm{|K|}$ biased categories occur exclusively and encourage context sharing in all other scenarios. 

\noindent \textbf{Implementation details:} For both proposed methods, we use ResNet-50~\cite{he-cvpr16} pre-trained on ImageNet as a backbone. For the first stage, an initial learning rate of $\mathrm{0.1}$ is used which is later divided by $\mathrm{10}$ following the standard step decay process for the learning rate. Following this, during the second stage of training, we train the network with a learning rate of $\mathrm{0.01}$ for both methods. For the CAM-based approach, we set $\mathrm{\lambda_{1}}$ and $\mathrm{\lambda_{2}}$ to be $0.1$ and $0.01$ respectively. 

The input images are randomly resize crop to $224 \times 224$ during the training. To further augment training data, we horizontal flip images. We use a batch size of $\mathrm{200}$ and stochastic gradient descent for optimization. Our model is implemented using PyTorch 1.0. Overall training time of both proposed methods is very close to that of a standard classifier and their inference time is exactly same as that of the standard classifier.
\section{Experiments}\label{sec:experiments}
In this section, we study the effectiveness of our approach across two tasks: object and attribute classification. We first describe our evaluation setup then report qualitative and quantitative performance on four image datasets against competitive baselines.
\begin{table}[t]
\centering
\footnotesize
\begin{tabular}{ | c | c | c | c | c |}
\hline
Datasets& Task & \#Classes & \#Train / \#Test \\ \hline
MS COCO + Stuff~\cite{caesar-cvpr18} & object & 171 & 82,783 / 40,504 \\ 
UnRel~\cite{peyre-iccv17} & object & 43 & - / 1,071 \\ \hline
Deep Fashion ~\cite{liu-cvpr16} & attribute &250 & 209,222/40,000\\ 
AwA~\cite{xian-tpami18}& attribute & 85 & 30,337 / 6,985\\ \hline
\end{tabular}
 \vspace{0.05in}
\caption{\footnotesize{\textbf{Properties of evaluation datasets.} For \cocoStuff, we use object training and validation data from COCO-2014 split~\cite{lin-eccv14}.}}
\vspace{-0.1in}
\label{table:dataset}
\end{table}

\begin{figure}[t!]
 \centering
 \includegraphics[width=0.25\textwidth]{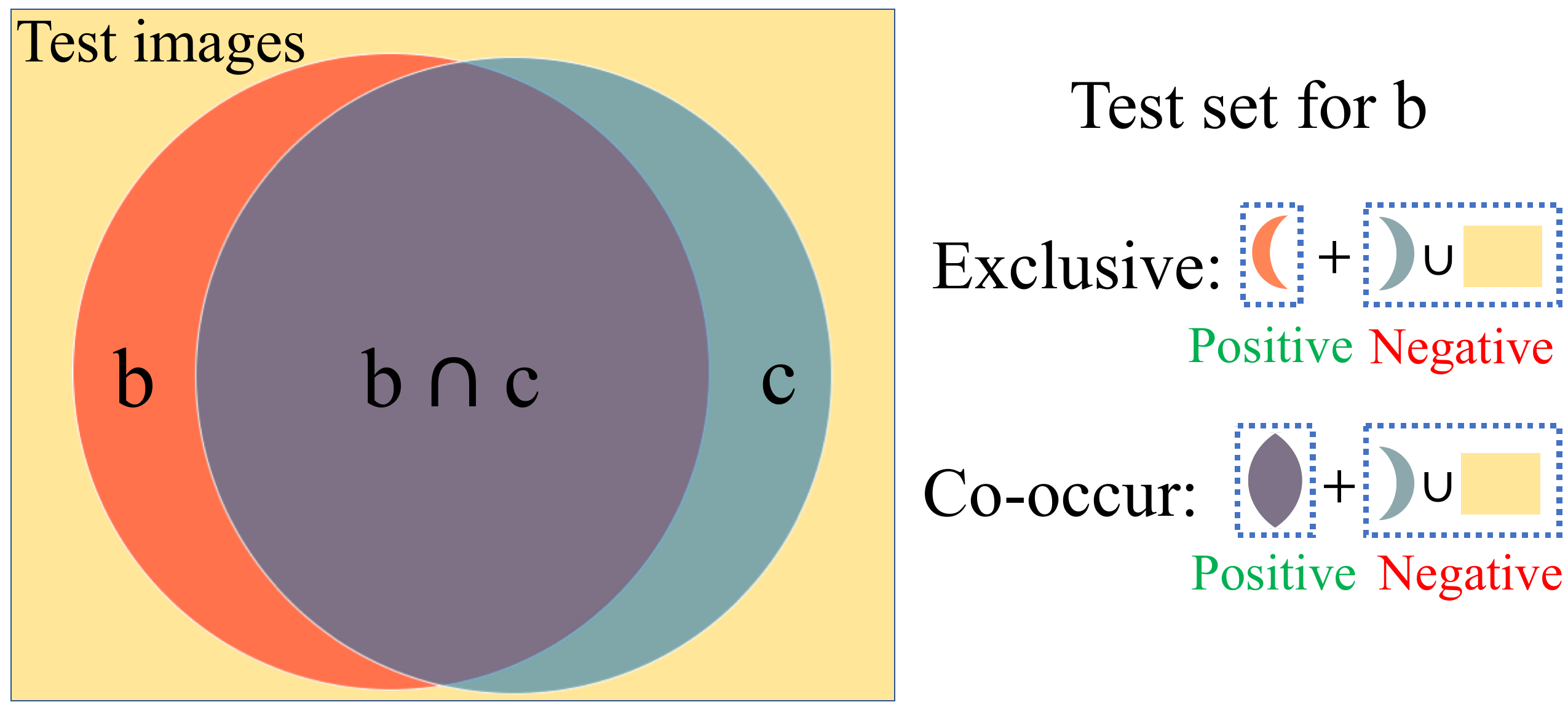}
 \caption{\footnotesize{\textbf{Our evaluation setup} has two different test data distributions: (1) \textbf{exclusive} and (2) \textbf{co-occurring}. Our goal is to improve recognition performance on (1) without compromising on (2).}}
 \vspace{-0.18in}
 \label{fig:test_split}
\end{figure}
\vspace{-0.2in}
\paragraph{Datasets:} We evaluate our approach on four multi-label datasets (summarized in Table~\ref{table:dataset}). The choice of these datasets was driven by the fact that they exhibit strong co-occurrence bias. We summarize their co-occurrence statistics in the supplementary material. For \deepFashion~\cite{liu-cvpr16}, we only consider $250$ most frequent attributes in the training data as other attributes do not have sufficient training samples. For \awa(AwA)~\cite{jayaraman-cvpr14,xian-tpami18}, following common practice, we train an attribute prediction network on seen ($\mathrm{40}$) animal categories and evaluate on unseen ($\mathrm{10}$) categories. Finally, \unrel dataset~\cite{peyre-iccv17} contains images of objects in unusual contexts, as they are obtained from rare and unusual triplet queries (e.g. ``person ride giraffe," ``dog ride bike''). We stress-test the generalizability of our model pre-trained on \cocoStuff on this dataset.
\vspace{-0.18in}
\paragraph{Evaluation setup:} We reiterate that our goal is to improve performance when highly biased categories occur exclusively, without losing much performance when they co-occur with other categories. Towards this end, for each dataset,
we first determine the most biased category pairs (\biasedSetPair) following the approach in Sec. \ref{sec:bias_formulation}. Next, for these $\mathrm{(b, c)}$ category pairs, we report performance on two different test data distributions: (1) \textbf{exclusive:} \biasedClass \textit{never} occurs with \cooccurClass and (2) \textbf{co-occur:} \biasedClass \textit{always} co-occurs with \cooccurClass. We illustrate the two test distributions in Fig.~\ref{fig:test_split}. We report top-3 recall for \deepFashion~\cite{liu-cvpr16} and mAP for all other datasets.
\vspace{-0.18in}
\paragraph{Baselines:}
Aside from a \emph{standard} classifier trained with a binary cross-entropy loss for each category, we compare with the following state-of-the-art methods that tackle the issue of co-occurring bias: (1) \textit{class balancing loss}~\cite{cui-cvpr19} by treating the scenarios where biased categories occur exclusively as tail classes and
(2) \dineshDecorrelate approach~\cite{jayaraman-cvpr14}, where we replace the hand-crafted features with deep network features (\texttt{conv5} features of ResNet-50) for a fairer comparison. 
To further test the strength of our method, we designed the following competitive baselines: 
\begin{packed_itemize}
\vspace{-0.05in}
\item \textit{remove co-occur labels}, where we remove labels corresponding to \cooccurClass for each \biasedClass in \biasedSetPair during training. By removing supervision about co-occurring categories, we intend to soften the context-induced bias on the model.
\item \textit{remove co-occur images} shares the same motivation as (2) but instead we remove training instances where the biased category and context co-occur.
\item \textit{weighted loss}, where we apply $10$ times higher weight to the loss when biased categories occur exclusively.
\item \textit{negative penalty}, where we assign a large negative penalty if the network predicts co-occurring category in cases where a biased category occurs exclusively. 
\end{packed_itemize}
\begin{table}[t]
\centering
\footnotesize
\begin{tabular}{ | c | c |c |}
\hline
Methods & Exclusive & Co-occur \\ \hline
\standard & 24.5 & \textbf{66.2} \\ 
\classBalance~\cite{cui-cvpr19} & 25.0 & 66.1 \\ \hline
\rmCoLabels & 25.2 & 65.9 \\ 
\rmCoImages & 28.4 & 28.7 \\ 
\wtLoss & \textbf{30.4} & 60.8 \\ 
\negPenalty & 23.8 &66.1 \\ \hline
\ourCam & 26.4 & 64.9 \\ 
\ourFeat & 28.8 & 66.0 \\ \hline
\end{tabular}
 \vspace{0.05in}
\caption{\footnotesize{\textbf{Performance on \cocoStuff} for the $20$ most biased categories. Both our methods perform very well on all baselines except \wtLoss and \rmCoImages on the exclusive test split, while successfully maintaining performance on the co-occurring test split.}}
\vspace{-0.23in}
\label{table:mscoco_results}
\end{table}

\vspace{-0.07in}
\subsection{Object Classification Performance} \label{sec:object_class}
\subsubsection{Overall Results}
In Table~\ref{table:mscoco_results}, we report performance on \cocoStuff for the $\mathrm{20}$ most biased categories. First, we observe that the \standard classifier has much better performance for co-occurring 
compared to exclusive test splits. This clearly demonstrates the inherent contextual bias present in \cocoStuff, as \standard classifier struggles when biased categories do not co-occur with context. \classBalance yields marginal gains indicating that weighing the rare exclusive cases alone cannot address contextual bias.

Next, we observe that both \ourCam
and \ourFeat outperform \standard by $\mathbf{1.9\%}$ and $\mathbf{4.3\%}$ respectively on the exclusive test set. \ourFeat has a very marginal drop of $0.2\%$ on the co-occurring split, compared to \standard, while the performance drop is higher for \ourCam. On categories such as ``ski'' and ``skateboard'' which have a very high co-occurrence bias with ``person'', the mAP boost from \ourFeat is $\mathbf{24.2\%}$ and $\mathbf{19.5\%}$ respectively (per-class mAP for both methods in supp. material).

\noindent \textbf{Comparison with other baselines:} We note that \rmCoImages approach performs poorly as it relies only on the exclusive images of the biased categories and do not take advantage of the vast amount of co-occurring images which supply complementary visual information. \wtLoss improves performance on the exclusive test split compared to \ourFeat (30.4\% vs. 28.8\%), but significantly hurts performance on co-occurring split (60.8\% vs. 66.0\%). \negPenalty does not hurt co-occurring split, but has inferior performance compared to our methods on the exclusive split. We also note that performance trends exhibited by these methods are consistent across all other datasets we test on; for all future experiments, we compare our methods with \standard and \classBalance. 

\noindent \textbf{Performance on the non-biased categories:} We evaluate on the $60$ non-biased object categories of \cocoStuff and observe that both \ourCam and \ourFeat perform on par with \standard, with a very mild drop of $~0.2\%$ overall mAP (details in supp. material). This indicates that our methods, while successfully improving performance for the biased categories, do not adversely effect the rest of the (non-biased) categories.
\begin{figure}[t!]
 \centering
 \includegraphics[width=0.43\textwidth]{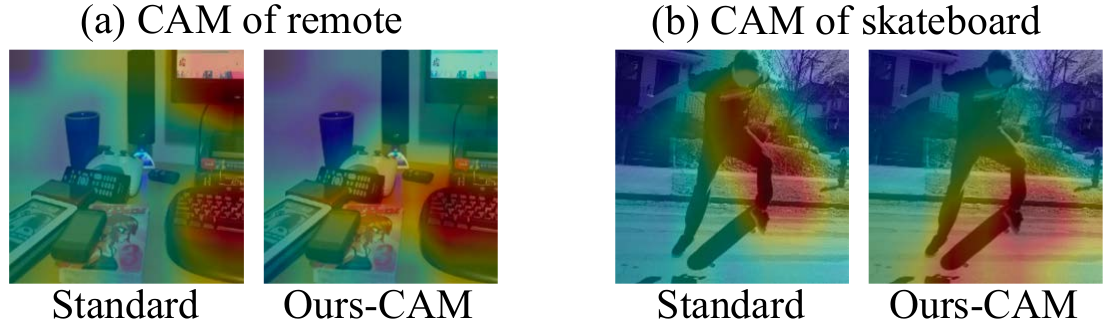}
 \caption{\footnotesize{\textbf{Learning from the right thing: \ourCam} (a) ``remote'' is contextually-biased by ``person.'' In the absence of ``person,'' \ourCam focuses on the right pixel regions compared to \standard. (b) ``skateboard'' co-occurs with ``person.'' \standard wrongly focuses on ``person'' due to contextual bias, while \ourCam rightly focuses on ``skateboard.''}}
 \label{fig:cam_success_failure}
 \vspace{-0.1in}
\end{figure}
\begin{figure}[t]
 \centering
 \includegraphics[width=0.45\textwidth]{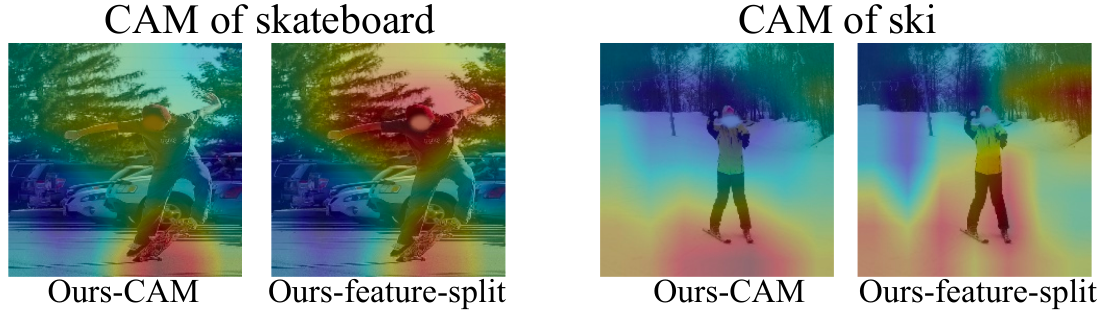}
 \caption{\footnotesize{\textbf{\ourCam vs. \ourFeat} on the images for which \ourFeat is able to recognize where as \ourCam fails. \ourCam primarily focuses on the object and does not use context whereas \ourFeat makes use of context for better prediction. 
}}
 \vspace{-0.12in}
 \label{fig:cam_feat}
\end{figure}

\begin{figure}[t!]
 \centering
 \includegraphics[width=0.3\textwidth]{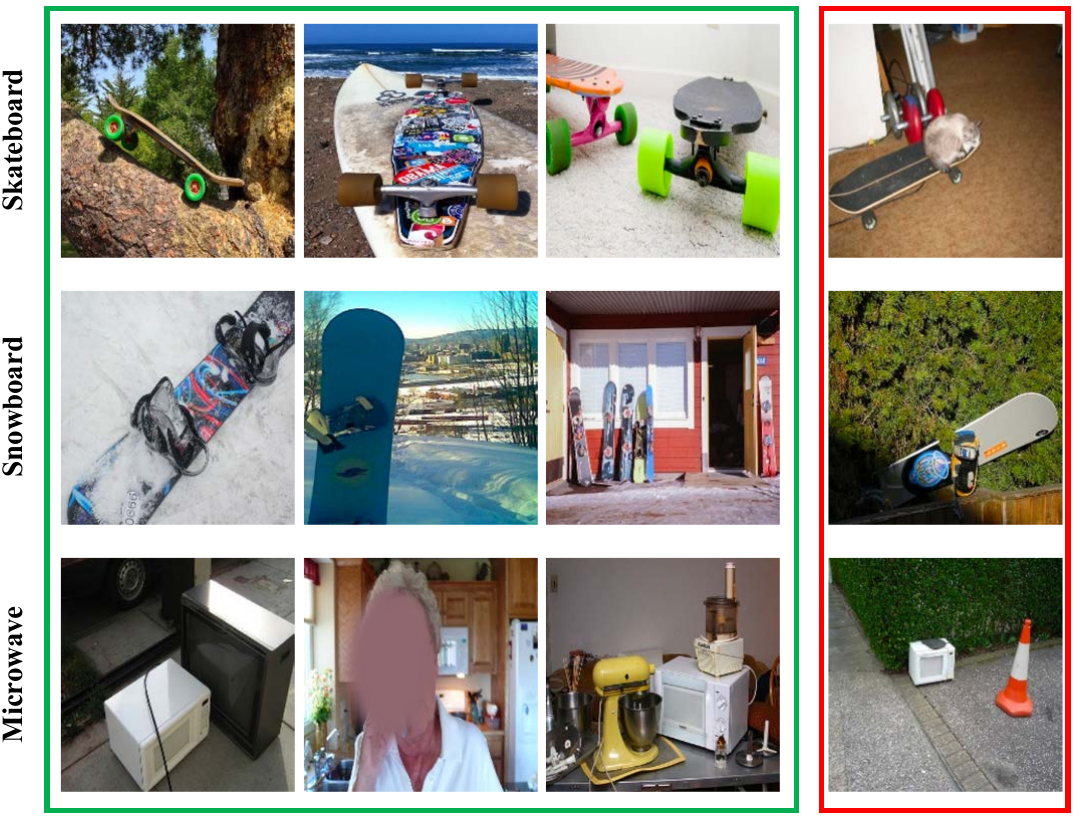}
 \caption{\footnotesize{\textbf{Learning from the right thing: \ourFeat} First $3$ columns indicate \textbf{success cases} where \ourFeat recognizes biased categories occurring away from their context while \standard fails. Last column: \textbf{failure cases} where both \standard and \ourFeat fail.}}
 \vspace{-0.18in}
 \label{fig:contex-result}
\end{figure}
\vspace{-0.1in}
\subsubsection{Qualitative Analysis}

\begin{figure}[t!]
 \centering
\vspace{-0.12in}
 \includegraphics[width=0.35\textwidth]{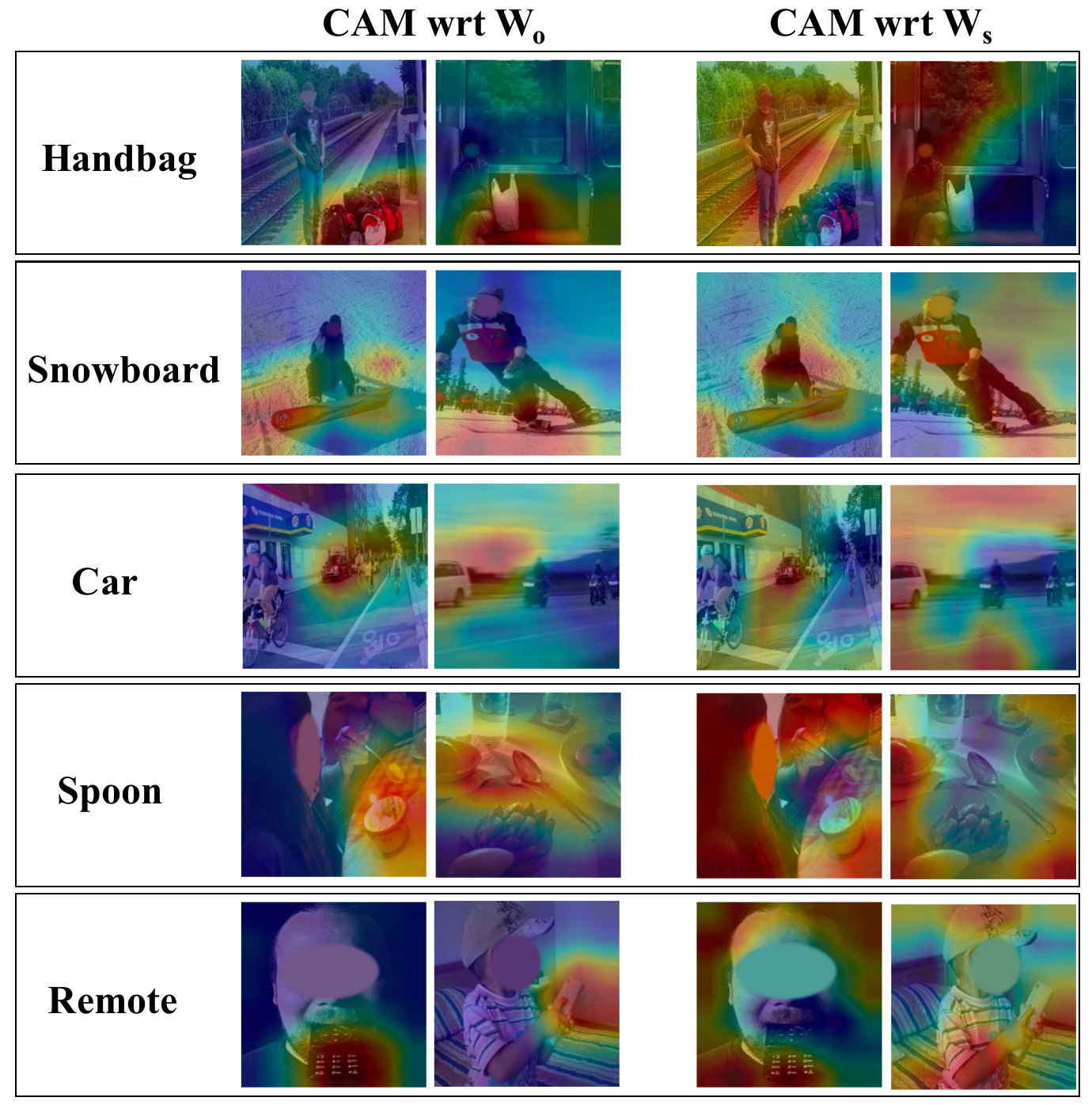}
 \caption{\footnotesize{\textbf{Interpreting \ourFeat} by visualizing CAMs with respect to \objectWeight (left) and \contextWeight (right). \objectWeight has learnt to consistently focus on the actual category (e.g., car) while \contextWeight captures context (e.g., road).}}
 \label{fig:main-result}
 \vspace{-0.1in}
\end{figure}

Next, we use CAM as a visualizing tool to analyze how our methods effectively tackle contextual bias.
\\
\noindent\textbf{\standard vs. \ourCam:} In Fig.~\ref{fig:cam_success_failure}, we present evidence where \standard fails but \ourCam succeeds\footnote{We determine `success' when the predicted probability is $ >= 0.5$ and `failure' otherwise.} to recognize biased categories. 
In both cases where a biased category co-occurs with context as well as occurs in its absence, \ourCam focuses on the right category thus ``learns from the right thing.''
\\ \noindent\textbf{\ourCam vs. \ourFeat:}
Fig.~\ref{fig:cam_feat} presents cases where \ourFeat succeeds but \ourCam struggles to recognize biased categories. We observe that while \ourCam rightly focuses on the category's pixel regions, \ourFeat additionally leverages the available context and thus performs better. 
\\\noindent\textbf{\standard vs. \ourFeat:}
The first $3$ columns in Fig.~\ref{fig:contex-result} present evidence where the \standard classifier fails but \ourFeat succeeds. For example, our method is able to recognize ``skateboard'' and ``snowboard'' in the absence of ``person'', and ``microwave'' in the absence of ``oven''. By contrast, the \standard classifier relies more on the context, thus fails on these images. The last column presents some failure cases where both \ourFeat and \standard  fail when biased categories occur without context. Common failure cases are challenging scenarios when the image has poor lighting, the object is zoomed out and thus very small (e.g., microwave).
\\ \noindent \textbf{Analysing \objectWeight and \contextWeight:}
Recall that in Sec.~\ref{sec:framework}, \ourFeat is formulated with a goal to prominently capture biased category-specific features through \objectWeight and context through \contextWeight. We visually verify this by generating two distinct class activation maps:
(i) \biasedFeat weighted by \objectWeight and (ii) \cooccurFeat weighted by \contextWeight. From Fig.~\ref{fig:main-result}, it is evident that \objectWeight learns to prominently focus on the category (e.g., handbag, car) and \contextWeight on the co-occurring context (e.g., person, road).

\begin{table}[t]
\vspace{0.1in}
\centering
\footnotesize
\begin{tabular}{ | c | c | c | c |}
\hline
Methods & \standard & \ourCam & \ourFeat \\ \hline
mAP & 42.0 & 45.3 & \textbf{52.1} \\ \hline
\end{tabular}
 \vspace{0.1in}
\caption{\footnotesize{\textbf{Cross-dataset experiment} where models trained on \cocoStuff are applied without fine-tuning on \unrel. \ourFeat yields huge boost over \standard highlighting its generalizability on unseen data.}}
\vspace{-0.2in}
\label{table:unrel_results}
\end{table}

\subsection{Cross dataset experiment on UnRel}
\vspace{-0.08in}
We next perform a \emph{cross-dataset} experiment by taking our models trained on \cocoStuff and testing them directly --- without any fine-tuning --- on \unrel dataset. \unrel has objects that are out-of-context (e.g., cat on a skateboard). Thus, a model that truly understands what the object is would be able to correctly classify it compared to a model that relies heavily on (or confuses the object with) context. Thus, this setting is a great testbed to evaluate our methods. Because we do not finetune, we evaluate only on the $3$ categories of \unrel that overlap with the $\mathrm{20}$ biased categories of \cocoStuff. From Table~\ref{table:unrel_results}, we observe that both \ourCam and \ourFeat outperform \standard by a large margins. This clearly demonstrates that both our methods learn from the right category and overcome contextual bias. 

\subsection{Attribute Classification}
\begin{table}[t!]
    \begin{center}
        \hspace*{-5pt}
		\tabcolsep=0.1cm
		\scriptsize
		\resizebox{0.49\textwidth}{!}{
		\begin{tabular}{| c | c  c | c  c  | }
		\hline
		    & \multicolumn{2}{| c |}{\begin{tabular}{@{}c@{}}\deepFashion \\ (top-3 recall)\end{tabular}} & \multicolumn{2}{| c |}{\begin{tabular}{@{}c@{}}\awa \\ (mAP)\end{tabular}}\\
			\hline
			Methods & Exclusive & Co-occur & Exclusive & Co-occur\\
			\hline
			\standard &  4.9 & 17.8 &  19.4 &  72.2 \\
			\classBalance~\cite{cui-cvpr19}  &  5.2 & 19.4 & 20.4 & 68.4  \\
			\dineshDecorrelate~\cite{jayaraman-cvpr14} &  - & - &  18.4 &  70.2\\ \hline
			\ourFeat &  \textbf{9.2} & \textbf{20.1} &  \textbf{20.8} &  \textbf{72.8}   \\
		
			\hline
		\end{tabular}}
		\vspace{0.01in}
		\caption{\footnotesize{\textbf{Attribute Classification Performance:} on \deepFashion and \awa computed on the $20$ most biased attributes. \ourFeat offers boosts over all approaches for the exclusive test split, without hurting performance on the co-occurring split.}}
		\label{table:attribute_res}
		\vspace{-0.35in}
	\end{center}
\end{table}

\vspace{-0.08in}
Here, we show that our approach of reducing contextual bias generalizes to attributes.
Our CAM-based approach is not applicable to attributes, as they lack well-defined spatial extents (details in Sec.~\ref{sec:cam}). As noted in Sec~\ref{sec:object_class}, the inherent contextual bias and difficulty in recognizing biased categories in the absence of their context leads to low scores on exclusive test split for all methods and datasets.

\noindent\textbf{Results on \deepFashion:}
As is the common practice, we report per class top-$3$ recall on \deepFashion~\cite{liu-cvpr16}. From Table~\ref{table:attribute_res}, we note that \ourFeat outperforms \standard by a significant margin on both test splits. For attributes like \textit{trapeze} and \textit{bell} which exhibit strong co-occurrence with \textit{striped} and \textit{lace} respectively, \ourFeat yields a boost of $\mathbf{21.2\%}$ and $\mathbf{17.4\%}$ top-3 recall respectively compared to \standard classifier. We present per-attribute results and comparisons with other baselines in the suppl. material.

\noindent\textbf{Results on \awa:} \awa~\cite{xian-tpami18} suffers from severe bias among attributes, e.g. \emph{blue} and \emph{spots} are highly correlated to \emph{coastal} and \emph{long leg} respectively. In this task, the goal is to learn an attribute classifier on ``seen'' animal categories (e.g ``spots'' attribute from the animal category ``dalmatian'') and evaluate the model's generalizability on \textit{unseen} animal categories (e.g. ``spots'' attribute on the unseen animal category ``leopard''). From Table \ref{table:attribute_res}, we observe that \ourFeat offers gains on the exclusive test split over other methods without hurting the co-occurring case. In particular, we outperform \dineshDecorrelate~\cite{jayaraman-cvpr14}, which was specifically designed to decorrelate attributes. 
\vspace{-0.1in}
\section{Conclusion}
\vspace{-0.1in}

We demonstrated the problem of contextual bias in popular object and attribute datasets by showing that standard classifiers perform poorly when biased categories occur away from their typical context. To tackle this issue, we proposed two simple yet effective methods to decorrelate feature representations of a biased category from its context. Both methods perform better at recognizing biased classes occurring away from their co-occurring context while maintaining the overall performance. More importantly, our methods generalize to new unseen datasets and perform significantly better than standard methods. Our current framework tackles contextual bias between pairs of categories; future efforts should leverage more available (scene or category) information and model relationships between them. Extending proposed methods to tasks like object detection and video action recognition is a worthy future direction.
\paragraph{Acknowledgments.} {This work was supported in part by NSF CAREER IIS-1751206.}

{\small
\bibliographystyle{ieee_fullname}
\bibliography{egbib}
}
\clearpage

\section*{Appendix}

\section{Additional implementation Details}
\noindent \noindent \textbf{Choosing the biased category pairs:} As mentioned in Sec. 3.1, our method is built on the following intuition: a given category \biasedClass is most biased by \cooccurClass if (1) the prediction probability of \biasedClass drops significantly in the \textit{absence} of \cooccurClass and (2) \biasedClass co-occurs frequently with \cooccurClass. Regarding (2), the co-occurring class for the biased categories appeared at least 20\% of the times with the biased categories on \cocoStuff and \awa dataset, and 10\% of the times for the \deepFashion dataset. 

 For the \cocoStuff, we partition the training data into non-overlapping $80-20$ split. We train a standard multi-label classifier with BCE loss on the $\mathrm{80\%}$ split and compute $\mathrm{bias}$ (Eq. 1) on the $\mathrm{20\%}$ split. For the \deepFashion, we train the classifier on the entire training data and determine the bias on the validation data.  For the \awa dataset, we need to use the test data to determine the biased classes as the test set has different
distribution than the training data (test set consists of animal classes unseen during the training).

\textbf{\noindent Choice of \alphaMin:} \alphaMin is set to $3$ for \cocoStuff and \awa, whereas it is set to $5$ for \deepFashion dataset. We found these values through cross-validation. During inference, a single forward pass of an image takes $\mathrm{0.2}$ ms on a single Titan X GPU.

\section{More results}
\noindent \textbf{Another baseline \splitBias:} In addition to all baselines we describe in the main text, we also designed another baseline: \splitBias. For this, we split each \biasedClass into two categories: (1) \biasedClass $\mathrm{\setminus}$ \cooccurClass and (2) \biasedClass $\mathrm{\cap}$ \cooccurClass. This setup adds $\mathrm{K}$ additional categories to each dataset and explicitly separates the two scenarios (exclusive and co-occur) for biased categories. This baseline is similar to~\cite{sadeghi-cvpr11}, where a separate classifier is learned for a visual phrase consisting of objects associated with a relation (e.g. ``person riding horse"). Here, instead of visual phrases, we learn a separate classifier for each co-occurring biased class pair. 
\subsection{Object Classification}
\textbf{Comparison with \splitBias :} Results in Table~\ref{table:mscoco_split_bias} shows that \ourFeat outperforms \splitBias with a significant margin on \cocoStuff ($28.8$ vs. $19.1$). Also, \ourCam gives much better performance than \splitBias ($26.4$ vs. $19.1$). Given that \splitBias cannot take full advantage of the co-occurring images (and vice-versa), it has inferior performance compared to both our methods.

\begin{table}[!t]
\centering
\footnotesize
\begin{tabular}{ | c | c |c |}
\hline
Methods & Exclusive & Co-occur \\ \hline
\splitBias & 19.1 & 64.3 \\ 
\ourCam & 26.4 & 64.9 \\ 
\ourFeat & 28.8 & 66.0 \\ \hline
\end{tabular}
 \vspace{0.05in}
\caption{\footnotesize{\textbf{Performance on \cocoStuff} for the $20$ most biased categories. \ourCam and \ourFeat outperform \splitBias with significant margin on both exclusive and co-occurring images.}}
\vspace{-0.1in}
\label{table:mscoco_split_bias}
\end{table}

\begin{table}[t]
\footnotesize
\centering
\begin{tabular}{ | c | c | c |}
\hline
Methods & 60 non biased categories & 171 object + stuff  \\ \hline
\standard & 75.4  & 57.2\\ \hline
\ourCam & 75.2 & 57.0 \\ \hline
\ourFeat & 75.2 & 57.1 \\ \hline

\end{tabular}
 \vspace{0.05in}
\caption{\footnotesize{\textbf{mAP of the non-biased object classes} and entire object+ stuff classes. Our approach loses only negligible  mAP compared to \standard classifier in these cases. }}
\label{table:non_bias_cls_result}
\end{table}

\textbf{Performance on non-biased classes:} In Table~\ref{table:non_bias_cls_result}, we show the mAP of our approach and \standard classifier on the non-biased object classes ($60$ classes) and on the entire \cocoStuff dataset (\emph{object + stuff}, 171 classes).
We can see that our approach very marginally ($~0.02\%$) reduces the performance on non-biased object and stuff classes, while improving performance when biased categories occur away from their context.

\begin{table}[t]
\centering
\footnotesize
\begin{tabular}{ | c | c |}
\hline
Methods & Cosine-similarity \\ \hline
\standard & 0.21 \\ \hline
\ourCam & 0.19  \\ \hline
\ourFeat & 0.17  \\ \hline

\end{tabular}
 \vspace{0.05in}
\caption{\footnotesize{\textbf{Cosine similarity} between classifier weights of the biased class pairs (\biasedClass,\cooccurClass). Our approach reduces the similarity between them indicating the biased class \biasedClass is less dependent on \cooccurClass for prediction.}}
\label{table:cos_sim}
\end{table}

\textbf{Measuring cosine-similarity between \objectWeight and \contextWeight :} 

We verify that \objectWeight and \contextWeight capture distinct information by computing a cosine similarity metric between them. From Table~\ref{table:cos_sim}, we observe that both our approaches yield a lower similarity score compared to \standard.

\textbf{Per class mAP and co-occurrence bias for 20 biased classes:} In the Table~\ref{table:perclass_coco}, we show per class results for the \cocoStuff for the top 20 biased classes. We also show the co-occurrence bias value for each class computed according to Eq. 1 in the main paper. From these results, we may observe that when a category occurs out of its context \ourFeat gives better performance compared to \standard classifier while maintaining the performance when a category co-occurs with context. \ourCam performs better than \standard when a category occurs away from its context, but struggles when categories co-occur.

\begin{figure}[t!]
    \centering
    \includegraphics[width=0.45\textwidth]{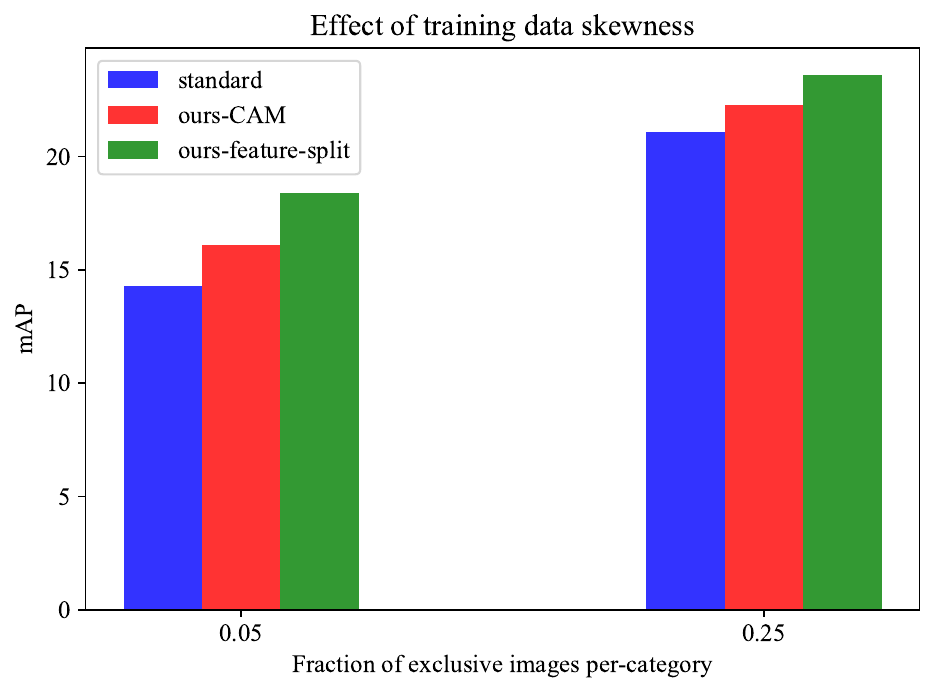}
    \caption{\footnotesize{mAP of  \standard , \ourCam , and \ourFeat classifier by varying fraction of exclusive images during training. If ratio is more skewed then we get a bigger boost for the exclusive cases. }}
    \label{fig:ex_co_ratio}
     \vspace{-0.2in}
\end{figure}
\textbf{Ablation study of \ourFeat by varying fraction of biased category images:} Here, we study the performance of our method as we vary the fraction of training images with biased categories occurring away from their typical context for \cocoStuff. Specifically, for each of the $20$ biased categories in \cocoStuff, we fix the total number of training images and vary the fraction of exclusive images. From Fig.~\ref{fig:ex_co_ratio}, we note that \standard performs rather poorly at lower fractions compared to both approaches (\ourCam and \ourFeat). Thus, both proposed methods achieve higher boosts at a fraction of $0.05$ compared to $0.25$. We also observe that a higher fraction of exclusive images benefits all the approaches, yet, our methods consistently outperform \standard. This indicates that our approaches are more robust than the baseline especially on heavily skewed training data.

\begin{table}[t]
\centering
\footnotesize
\begin{tabular}{ | c | c |c | }
\hline
Methods & Exclusive & Co-occur  \\ \hline
\standard & 4.9 & 17.8  \\ \hline
\splitBias & 3.5 & 14.3  \\ \hline
\rmCoLabels & 6.0 & \textbf{20.4}  \\ \hline
\rmCoImages & 4.2 & 5.4  \\ \hline
\negPenalty & 5.5 & 18.9  \\ \hline
\classBalance~\cite{cui-cvpr19} & 5.2 & 19.4  \\ \hline
\ourFeat  & \textbf{9.2} & 20.1 \\ \hline
\end{tabular}
\vspace{0.05in}
\caption{\footnotesize{\textbf{Top-3 recall on \deepFashion} for the $20$ most biased attributes. \ourFeat yields a significant boost over all approaches for the exclusive test split, without hurting performance on the co-occurring split. \ourCam is not extensible to attributes hence not reported here. The above baseline methods are described in our main paper.}}
\label{table:deepfashion_results}
\end{table}

\begin{table}[t]
\centering
\footnotesize
\begin{tabular}{ | c | c | c |}
\hline
Methods & Exclusive & Co-occur \\ \hline
\standard & 19.4 & 72.2 \\ \hline
 \splitBias & 19.7 & 66.8 \\ \hline
 \rmCoLabels & 19.1 & 62.9 \\ \hline
 \rmCoImages & \textbf{22.7} & 58.3 \\ \hline
 \negPenalty & 19.2 & 68.4 \\ \hline
\classBalance~\cite{cui-cvpr19} & 20.4 & 68.4 \\ \hline
\dineshDecorrelate~\cite{jayaraman-cvpr14} & 18.4 & 70.2 \\ \hline
\textbf{\ourFeat} & 20.8 & \textbf{72.8} \\ \hline
\end{tabular}
 \vspace{0.05in}
\caption{\footnotesize{\textbf{Performance on \awa} for the $20$ most biased attributes. Our proposed method \ourFeat outperforms other methods. \ourCam is not extensible to attributes hence not reported here.}}
\label{table:animal_results}
\end{table}

\subsection{Comparison with other baselines for attribute classification}

 Table~\ref{table:deepfashion_results} reports performance on \deepFashion~\cite{liu-cvpr16}.  We outperform all baselines by a significant margin on the exclusive test set. Although \rmCoLabels has slightly higher performance when attributes co-occur ($20.4$ vs. $20.1$), \ourFeat performs significantly better when attributes occur exclusively ($6.0$ vs. $9.2$).

 From Table \ref{table:animal_results}, we observe that \ourFeat offers gains on the exclusive test split compared to most methods for \awa dataset. Though \rmCoImages yields higher gains on the exclusive test split, unlike \ourFeat, it severely hurts the performance of co-occurring cases. Meanwhile \ourFeat achieves good gains in exclusive cases without hurting co-occurring cases.

Finally, in  Table~\ref{table:perclass_deepfashion} and \ref{table:perclass_awa}, we show per category performance for the top $20$ biased categories for two datasets: \deepFashion and \awa. These results show that \ourFeat gives better performance than the \standard classifier when attributes occur exclusively without their co-occurring context. At the same time, \ourFeat maintains performance when biased attribute categories appear with co-occurring context.

\begin{table*}[t]
\centering
\tabcolsep=0.1cm
\begin{tabular}{| c  c c | c  c  c| c  c  c| }
\hline
\multicolumn{3}{| c |}{Classes} & \multicolumn{3}{| c |}{Exclusive} & \multicolumn{3}{| c |}{Co-occur}\\
			\hline
Biased class & Co-occur class & Bias  & \standard & \ourCam & \ourFeat & \standard & \ourCam & \ourFeat \\ \hline
cup &	dining table &	1.76 &	33.0 & 	35.4 & 	27.4 & 	68.1 & 	63.0 & 	70.2 \\ 
wine glass &	person &	1.8	 & 35.0 & 	36.3 & 	35.1 & 	57.9 & 	57.4 & 	57.3 \\ 
handbag	& person &	1.81 &	3.8 & 	5.1 & 	4.0 & 	42.8 & 	41.4 & 	42.7 \\ 
apple &	fruit &	1.91 &	29.2 & 	29.8 & 	30.7 & 	64.7 & 	64.4 & 	64.1 \\ 
car	 & road &	1.94 &	36.7 & 	38.2 & 	36.6 & 	79.7 & 	78.5 & 	79.2 \\ 
bus &	road &	1.94 &	40.7 & 	41.6 & 	43.9 & 	86.0 & 	85.3 & 	85.4 \\ 
potted plant &	vase &	1.99 &	37.2 & 	37.8 & 	36.5 & 	50.0 & 	46.8 & 	46.0 \\ 
spoon &	bowl &	2.04 &	14.7 & 	16.3 & 	14.3 & 	42.7 & 	35.9 & 	42.6 \\ 
microwave &	oven &	2.08 &	35.3 & 	36.6 & 	39.1 & 	60.9 & 	60.1 & 	59.6 \\ 
keyboard &	mouse &	2.25 &	44.6 & 	42.9 & 	47.1 & 	85.0 & 	83.3 & 	85.1 \\ 
skis &	person &	2.28 &	2.8 & 	7.0 & 	27.0 & 	91.5 & 	91.3 & 	91.2 \\ 
clock &	building &	2.39 &	49.6 & 	50.5 & 	45.5 & 	84.5 & 	84.7 & 	86.4 \\ 
sports ball &	person &	2.45 &	12.1 & 	14.7 & 	22.5 & 	75.5 & 	75.3 & 	74.2 \\ 
remote &	person &	2.45 &	23.7 & 	26.9 & 	21.2 & 	70.5 & 	67.4 & 	72.7 \\ 
snowboard &	person &	2.86 &	2.1 & 	2.4 & 	6.5 & 	73.0 & 	72.7 & 	72.6 \\ 
toaster	 & ceiling &	3.7 &	7.6 & 	7.7 & 	6.4 & 	5.0 & 	5.0 & 	4.4 \\ 
hair drier	& towel &	4 &	1.5 & 	1.3 & 	1.7 & 	6.2 & 	6.2 & 	6.9 \\ 
tennis racket &	person &	4.15 &	53.5 & 	59.7 & 	61.7 & 	97.6 & 	97.5 & 	97.5 \\ 
skateboard &	person &	7.36 &	14.8 & 	22.6 & 	34.4 & 	91.3 & 	91.1 & 	90.8 \\ 
baseball glove &	person &	339.15 &	12.3 & 	14.4 & 	34.0 & 	91.0 & 	91.3 & 	91.1 \\ 
\hline
Mean &	- & - &	24.5 & 26.4 &	\textbf{28.8} &	\textbf{66.2} &	64.9 & 66.0 \\
\hline
\end{tabular}
\caption{\footnotesize{\cocoStuff dataset. Per class mAP and bias for 20 most biased classes. \ourFeat outperforms \standard on the exclusive set while maintaining the performance on the co-occurring cases.}}
\label{table:perclass_coco}
\end{table*}

\begin{table*}[t]
\centering
\begin{tabular}{| c  c c| c  c | c  c | }
\hline
\multicolumn{3}{| c |}{Classes} & \multicolumn{2}{| c |}{Exclusive} & \multicolumn{2}{| c |}{Co-occur}\\
			\hline
Biased class & Co-occur class & Bias  & \standard & \ourFeat & \standard & \ourFeat \\ \hline
bell &	lace &	3.15 &	5.4 & 	22.8 & 	3.1 & 	9.4 \\ 
cut	& bodycon &	3.3 &	8.6 & 	12.5 & 	29.3 & 	36.2 \\ 
animal	& print &	3.31 &	0.0 & 	1.9 & 	1.9 & 	2.8 \\ 
flare &	fit &	3.31 &	18.4 & 	32.0 & 	56.0 & 	62.0 \\ 
embroidery &	crochet &	3.44 &	4.1 & 	1.8 & 	4.8 & 	0.0 \\ 
suede &	fringe &	3.48 &	12.0 & 	19.6 & 	65.2 & 	73.9 \\ 
jacquard &	flare &	3.68 &	0.0 & 	0.9 & 	0.0 & 	9.1 \\ 
trapeze	& striped &	3.7	& 8.7 & 	29.9 & 	42.9 & 	50.0 \\ 
neckline &	sweetheart &	3.98 &	0.0 & 	0.0 & 	0.0 & 	0.0 \\ 
retro &	chiffon &	4.08 &	0.0 & 	0.4 & 	0.0 & 	0.0 \\ 
sweet &	crochet &	4.32 &	0.0 & 	0.5 & 	0.0 & 	0.0 \\ 
batwing	& loose &	4.36 &	11.0 & 	12.0 & 	27.5 & 	15.0 \\ 
tassel &	chiffon &	4.48 &	13.0 & 	16.8 & 	25.0 & 	25.0 \\ 
boyfriend &	distressed &	4.5 &	11.6 & 	11.6 & 	49.2 & 	38.1 \\ 
light &	skinny &	4.53 &	2.0 & 	1.3 & 	14.9 & 	8.5 \\ 
ankle &	skinny &	4.56 &	1.0 & 	14.6 & 	13.2 & 	27.9 \\ 
french &	terry &	5.09 &	0.0 & 	0.8 & 	9.6 & 	7.9 \\ 
dark &	wash &	5.13 &	2.6 & 	2.1 & 	8.7 & 	13.0 \\ 
medium &	wash &	7.45 &	0.0 & 	0.0 & 	0.0 & 	0.0 \\ 
studded	& denim &	7.8 &	0.0 & 	3.2 & 	4.0 & 	24.0 \\ 
\hline
Mean &	- & - &	4.9 & 	\textbf{9.2} & 	17.8 &	\textbf{20.1} \\
\hline
\end{tabular}
\caption{\footnotesize{\deepFashion dataset. Per class top-3 recall and bias for 20 most biased classes. \ourFeat outperforms \standard on the exclusive set while maintaining the performance on the co-occurring cases.}}
\label{table:perclass_deepfashion}
\end{table*}

\begin{table*}[t]
\centering
\begin{tabular}{| c  c c| c  c | c  c | }
\hline
\multicolumn{3}{| c |}{Classes} & \multicolumn{2}{| c |}{Exclusive} & \multicolumn{2}{| c |}{Co-occur}\\
			\hline
Biased class & Co-occur class & Bias  & \standard & \ourFeat & \standard & \ourFeat \\ \hline
white &	ground  &	3.67 &	24.8 & 	24.6 & 	85.8 & 	86.2 \\ 
longleg	 & domestic  &	3.71 &	18.5 & 	29.1 & 	89.4 & 	89.3 \\ 
forager	 & nestspot  &	4.02 &	33.6 & 	33.4 & 	96.6 & 	96.5 \\ 
lean  &	stalker  &	4.46 &	11.5 & 	12.0 & 	54.5 & 	55.8 \\ 
fish  &	timid  &	5.14 &	60.2 & 	57.4 & 	98.3 & 	98.3 \\ 
hunter  &	big  &	5.34 &	4.1 & 	3.6 & 	32.9 & 	30.0 \\ 
plains  &	stalker  &	5.4 &	6.4 & 	6.0 & 	44.7 & 	59.9 \\ 
nocturnal  &	white  &	5.84 &	13.3 & 	13.1 & 	71.2 & 	60.5 \\ 
nestspot  &	meatteeth  &	5.92 &	13.4 & 	14.9 & 	62.8 & 	67.6 \\ 
jungle  &	muscle  &	6.26 &	33.3 & 	31.3 & 	88.6 & 	86.6 \\ 
muscle	 & black  &	6.39 &	9.3 & 	9.3 & 	76.6 & 	73.6 \\ 
meat  &	fish  &	7.12 &	4.5 & 	3.8 & 	76.1 & 	73.6 \\ 
mountains  &	paws  &	9.24 &	10.9 & 	10.0 & 	49.9 & 	39.9 \\ 
tree  &	tail  &	10.98 &	36.5 & 	55.0 & 	93.2 & 	92.7 \\ 
domestic  &	inactive  &	11.77 &	11.9 & 	13.1 & 	73.7 & 	76.6 \\ 
spots  &	longleg  &	20.15 &	43.8 & 	45.2 & 	61.8 & 	59.1 \\ 
bush  &	meat  &	29.47 &	19.8 & 	22.1 & 	70.2 & 	75.1 \\ 
buckteeth  &	smelly  &	34.01 &	7.8 & 	8.9 & 	27.1 & 	45.3 \\ 
slow  &	strong  &	76.59 &	15.5 & 	14.6 & 	95.8 & 	93.3 \\ 
blue  &	coastal  &	319.98	& 8.4 & 	8.2 & 	94.2 & 	95.8 \\ 
\hline
Mean &	- &	- & 19.4 &	\textbf{20.8} &	72.2 &	\textbf{72.8} \\
\hline
\end{tabular}
\caption{\footnotesize{\awa dataset. Per class mAP and bias for 20 most biased classes. \ourFeat outperforms \standard on the exclusive set while maintaining the performance on the co-occurring cases.}}
\label{table:perclass_awa}
\end{table*}

\end{document}